\documentclass[11pt]{article}

\usepackage[final]{acl}

\usepackage{times}
\usepackage{latexsym}
\usepackage[T1]{fontenc}
\usepackage[utf8]{inputenc}
\usepackage{microtype}
\usepackage{graphicx}
\usepackage{amsmath}
\usepackage{amssymb}
\usepackage{booktabs}
\usepackage{multirow}
\usepackage{array}
\usepackage{xcolor}
\usepackage{url}
\usepackage{enumitem}
\usepackage{subcaption}

\definecolor{gain}{RGB}{0,110,0}
\definecolor{loss}{RGB}{180,0,0}

\title{TriLens: Per-Layer Logit-Lens Entropy for White-Box Hallucination Detection}

\author{
\textbf{Bohan Yang\textsuperscript{1,2},}
~\textbf{Yijun Gong\textsuperscript{5},}
~\textbf{Zhi Zhang\textsuperscript{1},}
~\textbf{Ge Zhang\textsuperscript{3},} \\
\textbf{Wenpeng Xing\textsuperscript{1,3},}
~\textbf{Meng Han\textsuperscript{1,3,4}}
\\
\textsuperscript{1}Binjiang Institute of Zhejiang University, \\
\textsuperscript{2}Beijing Normal-Hong Kong Baptist University, \\
~\textsuperscript{3}Zhejiang University,
~\textsuperscript{4}GenTel.io,
~\textsuperscript{5}Great Bay University \\
\fontsize{10.2pt}{0.1\baselineskip}\texttt{t330016056@mail.bnbu.edu.cn, \{wpxing, mhan\}@zju.edu.cn}
}

\begin{document}
\maketitle

\begin{abstract}
  When a language model hallucinates, the final answer is wrong, but the
  mistake is not necessarily invisible inside the model. Different internal
  pathways may remain uncertain, disagree in how quickly they sharpen, or
  commit to competing continuations before the output is produced. We
  introduce \emph{TriLens}, a white-box detector that turns this intuition
  into a compact representation: at every layer, it reads the multi-head
  self-attention output, the feed-forward output, and the residual stream
  through the model's own logit lens, then records only the entropy of each
  readout. The resulting $3L$-dimensional trajectory describes how certainty
  forms across depth and across modules, without storing high-dimensional
  hidden states or sampling multiple generations. This simple signal yields
  a strong detector across instruction-tuned LLMs and QA benchmarks, and
  our analyses show that the three module-wise entropy trajectories provide
  complementary evidence. TriLens suggests that hallucination detection can
  benefit from tracking how internal computation settles, not only what the
  final layer predicts. The code is available at
  \url{https://tosakaucw.github.io/TriLens/}.
\end{abstract}

\section{Introduction}
\label{sec:intro}

Large language models (LLMs) have reshaped a wide range of NLP tasks,
but their tendency to generate hallucinations---outputs that are fluent
yet factually wrong or inconsistent with provided
context---remains a fundamental obstacle to deployment in
knowledge-sensitive settings~\citep{ji2023survey,li2023halueval}.
Reliable \emph{detection} enables selective abstention, confidence
calibration, and targeted retrieval.

Among existing detection approaches, \emph{white-box} methods that
analyze an LLM's internal state during a single forward pass are
particularly attractive: they avoid the need for multiple samples or
external references while enabling efficient inference-time scoring.
Recent work has shown that hallucination-relevant information can be
extracted from hidden states, gradients, semantic-entropy probes,
spectral summaries, attention kernels, and update-consistency
features~\citep{azaria2023saplma,kossen2024sep,hu2024embedding,chen2024inside,sriramanan2024llmcheck,zhang2025icrprobe}.
However, these approaches still largely treat internal computation as a
high-dimensional representation to be classified, rather than asking
which low-dimensional aspect of the computation changes when a model
moves toward an unsupported answer.

\begin{figure}[t]
  \centering
  \includegraphics[width=\linewidth]{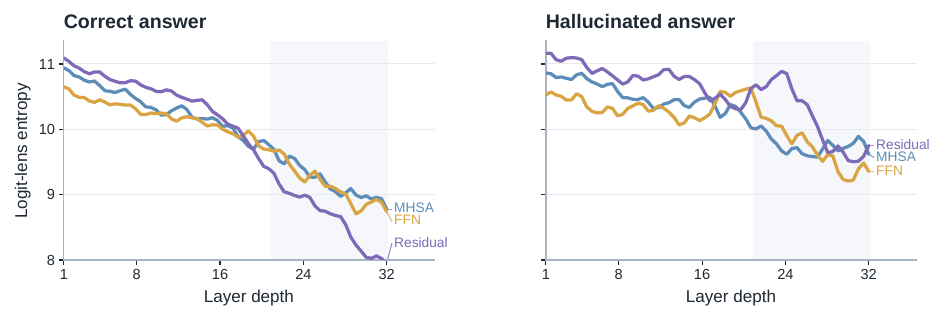}
  \caption{\textbf{TriLens: Per-layer logit-lens entropy tracks internal certainty during generation.}
    Supported answers show coordinated entropy sharpening across MHSA, FFN,
    and residual-stream readouts, whereas hallucinated answers tend to retain
    higher, less stable, and less synchronized entropy across depth.}
  \label{fig:teaser}
\end{figure}

This leads to two empirical questions. First, can hallucination be
detected from how internal certainty forms across depth, rather than from
final-layer confidence or high-dimensional hidden states? Second, do the
main transformer pathways provide redundant views of that uncertainty, or
do attention, feed-forward, and residual-stream readouts carry
complementary evidence? A useful detector should answer these questions
under a single forward pass, expose where uncertainty enters the
computation, and avoid turning every hidden state into a large learned
representation. We therefore test a simpler alternative: instead of
storing activations themselves, we ask how uncertain different parts of
the computation look when read through the model's own vocabulary lens.

\emph{TriLens} records one scalar from each of three internal locations at
every layer: the multi-head self-attention output, the feed-forward
output, and the residual stream. Each location is projected through the
logit lens, converted to a Shannon entropy, and concatenated across
depth. The result is a compact $3L$-dimensional trajectory that tracks
how certainty forms across pathways. It is small enough for a simple
probe, but structured enough to distinguish whether contextual routing,
parametric recall, and the composed residual state sharpen together or
remain diffuse.

\paragraph{Why should a first-order statistic suffice?}
Our working hypothesis is that hallucination-relevant errors are often
accompanied by elevated uncertainty across internal pathways of an LLM,
and that per-layer Shannon entropy of logit-lens distributions is a
compact diagnostic correlate of this effect. In a decoder
transformer, MHSA routes contextual information while the FFN pathway
retrieves parametric content~\citep{geva2021ffn,geva2023dissecting,elhage2021mathematical}.
When these pathways align, their logit-lens distributions typically
sharpen with depth; when they diverge, refinement can become slower and
higher-entropy. TriLens tracks this behavior with three scalars per
layer, yielding a compact alternative to high-dimensional hidden-state
features.

\paragraph{Contributions.}
Our main contribution is to test these questions through a compact
pathway-wise entropy feature, supported by ablations and diagnostic
comparisons.
\begin{itemize}[leftmargin=*,itemsep=1pt,topsep=2pt]
  \item We introduce \textbf{TriLens}, a per-layer logit-lens entropy
        feature over MHSA, FFN, and residual-stream states.
  \item Across three instruction-tuned LLMs and four benchmark settings,
        this $3L$-dimensional feature improves over the strong prior
        white-box baseline ICR Probe on \emph{all 12} cells, averaging
        $+12.1$ AUROC, and attaining the highest AUROC in 11 of the 12 cells.
  \item Ablations show that MHSA, FFN, and residual-stream entropies
        contribute complementary signal, while an intra-layer
        module-disagreement term adds little.
  \item A head-to-head comparison with DoLa-style cross-layer contrast
        features shows that per-layer entropy captures much of the useful
        signal in that family.
  \item Layer-wise analysis shows that the peak-discriminative depth
        varies systematically across architectures and benchmarks, consistent
        with differences in how models integrate contextual and parametric
        information.
\end{itemize}

\begin{figure*}[t]
  \centering
  \includegraphics[width=1.0\textwidth]{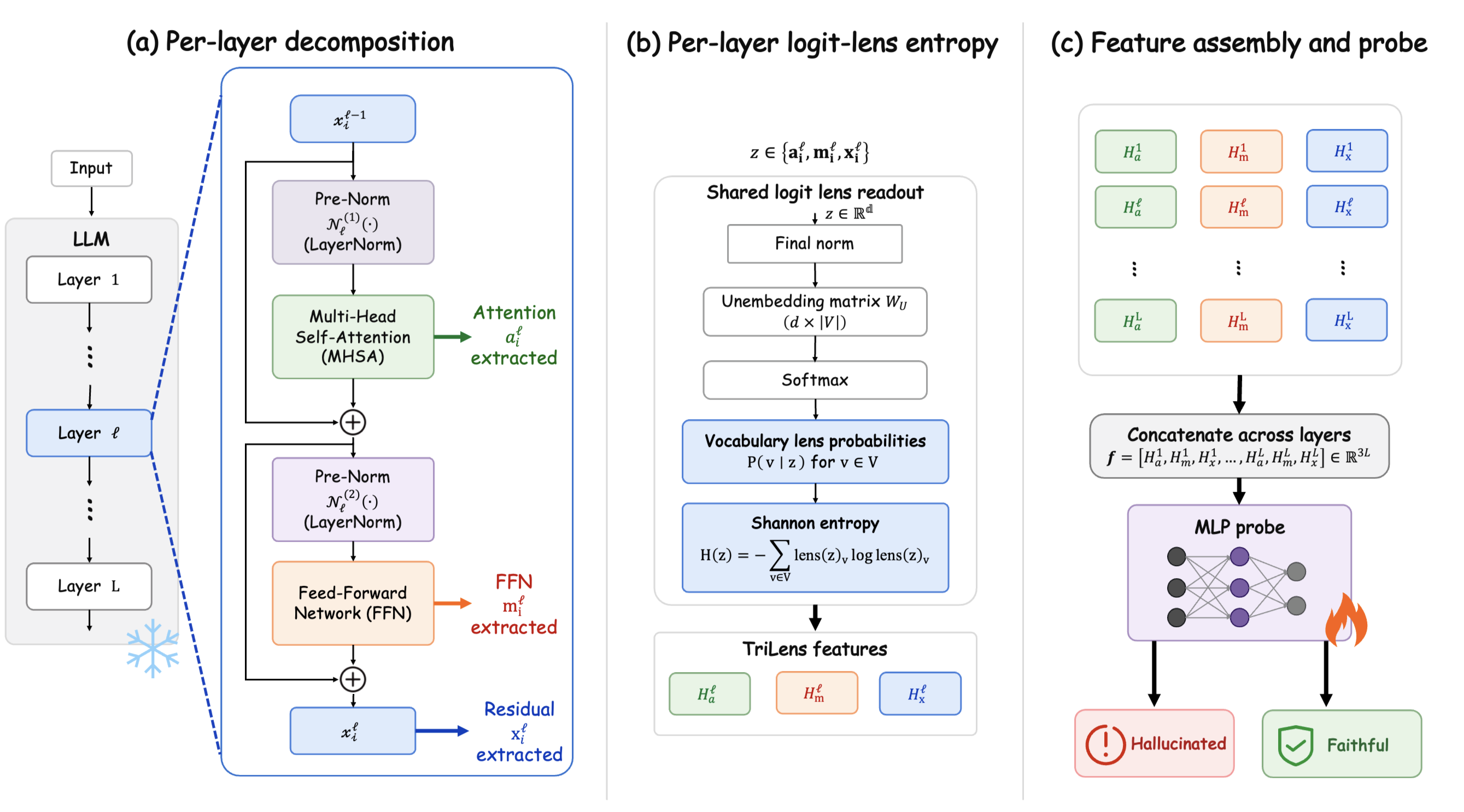}
  \caption{TriLens overview. A single forward pass over the evaluation
    sequence extracts the MHSA output, FFN output, and residual stream at every
    layer; each is projected through the model's logit lens, converted to a
    Shannon entropy, concatenated into a $3L$-dimensional feature, and fed to a
    probe.}
  \label{fig:trilens-overview}
\end{figure*}

\section{Related Work}
\label{sec:related}

\paragraph{White-box hallucination detection.}
The dominant white-box line trains detectors on internal activations
from a single forward pass.
SAPLMA~\citep{azaria2023saplma}, layer-combination
probes~\citep{chwang2024androids}, SEP~\citep{kossen2024sep},
gradient-aware detectors~\citep{hu2024embedding}, INSIDE~\citep{chen2024inside},
and LLM-Check~\citep{sriramanan2024llmcheck} all fit this template,
while ICR Probe~\citep{zhang2025icrprobe} reframes detection around
residual-stream updates. TriLens differs by using a compact $3L$-scalar
feature built from pathway-specific logit-lens entropy rather than
high-dimensional hidden states or update-consistency scores.

\paragraph{Uncertainty-based detection.}
A parallel black-box line scores hallucination risk from a model's own
uncertainty, including calibration-style signals~\citep{kadavath2022know,malinin2020entropy},
semantic uncertainty and semantic entropy~\citep{kuhn2023semantic,farquhar2024semantic},
and recent refinements based on Bayesian estimation, pairwise semantic
similarity, fact-level self-consistency, or log-probability time
series~\citep{ciosek2025budget,nguyen2025beyond,sawczyn2025factselfcheck,shapiro2026halt}.
TriLens instead uses a single white-box pass over internal activations,
without multiple sampled generations or output-layer uncertainty alone.

\paragraph{Logit-lens and mechanistic motivation.}
The logit lens and Tuned Lens project intermediate states into
vocabulary space~\citep{nostalgebraist2020logitlens,belrose2023tunedlens};
DoLa~\citep{chuang2024dola} contrasts these distributions across
layers. Our design is further motivated by mechanistic views of FFNs as
key-value memory, MHSA as residual-stream routing, and depth-varying
module influence~\citep{geva2021ffn,geva2023dissecting,elhage2021mathematical,stolfo2023mechanistic}.
TriLens combines these ingredients into a supervised detection feature
based on pathway-specific entropy trajectories; \S\ref{sec:dola}
compares it directly with DoLa-style cross-layer contrast.

\section{Method}
\label{sec:method}

\subsection{Per-Module Logit-Lens Entropy}
\label{sec:method-entropy}

We consider an LLM with $L$ decoder layers and vocabulary $V$.
At layer $\ell$ and token position $i$, the residual stream updates as
\begin{equation}
  \label{eq:update}
  \begin{aligned}
    x_i^\ell = x_i^{\ell-1}
     & + \underbrace{\text{MHSA}^\ell(\mathcal{N}_\ell(x_i^{\ell-1}))}_{a_i^\ell}          \\
     & + \underbrace{\text{FFN}^\ell(\mathcal{N}_\ell(x_i^{\ell-1}+a_i^\ell))}_{m_i^\ell}.
  \end{aligned}
\end{equation}
For a hidden activation $z \in \{x_i^\ell, a_i^\ell, m_i^\ell\}$, the
\emph{logit lens}~\citep{nostalgebraist2020logitlens} projects $z$ into
vocabulary-space probabilities
\begin{equation}
  \label{eq:entropy}
  \begin{aligned}
    \mathrm{lens}(z) & = \mathrm{softmax}\big(W_U\,\mathrm{Norm}_{\mathrm{final}}(z)\big), \\
    H(z)             & = -\sum_{v \in V} \mathrm{lens}(z)_v \log \mathrm{lens}(z)_v .
  \end{aligned}
\end{equation}
using the model's own unembedding matrix $W_U$ and final normalization layer.
Here $\mathcal{N}_\ell$ denotes the layer's pre-submodule normalization
(e.g., RMSNorm in the model families studied here), while the lens
itself always uses the model's \emph{final} normalization layer before
unembedding. This mirrors the standard logit-lens construction on the
residual stream and keeps $a^\ell$, $m^\ell$, and $x^\ell$ in the same
readout coordinate. In implementation we apply the final normalization
directly to each of $a^\ell$, $m^\ell$, and $x^\ell$, with no additional
centering or rescaling.
The \emph{per-layer logit-lens entropy} is the Shannon entropy of this
distribution; we compute it at all three positions, yielding
$H_a^\ell, H_m^\ell, H_x^\ell$ for every layer
(Eq.~\ref{eq:entropy}).
TriLens applies the same readout not only to the residual stream but
also to the isolated MHSA and FFN writes, using the resulting
three-pathway trajectory as a detection feature.

\subsection{Mechanistic Motivation}
\label{sec:method-mechanism}

The choice to measure entropy \emph{per module} rather than only on
$x^\ell$, and to use \emph{Shannon entropy} rather than richer summary
statistics, is not arbitrary. It follows from three established views
of transformer dynamics.

\paragraph{Dual-pathway decomposition.}
Under the residual-stream framework
of~\citet{elhage2021mathematical}, $a^\ell$ and $m^\ell$ are functionally
distinct pathways: the OV circuit of MHSA copies content from earlier
positions, while FFN functions as a read-out from a parametric
key-value memory~\citep{geva2021ffn,geva2023dissecting}.
In factual completion, both pathways often converge on the same
target token---MHSA because the context uniquely determines the
answer, FFN because the parametric memory has stored the relevant
key--value pair.
In hallucinated completion under our benchmark setting, the pathways can
\emph{diverge}: $a^\ell$ may still route contextually correct
information, while $m^\ell$ can partially recall the same correct
alternative even though the observed token is wrong.
Projecting $a^\ell$ and $m^\ell$ independently through the logit lens
exposes this divergence, which a single measurement on the composed
$x^\ell$ can mask whenever the two pathways cancel in direction but
not in magnitude.
Although $a^\ell$ and $m^\ell$ are not themselves full residual states,
they are additive writes in the same residual space and are therefore
legitimate objects for the model's own final readout. The resulting
distributions should be interpreted as \emph{readout preferences of an
  isolated write} rather than as standalone next-token predictions; our
claim is empirical and diagnostic, not that $a^\ell$ or $m^\ell$ alone
constitute a full decoder state. Accordingly, the role of $H_a^\ell$
and $H_m^\ell$ in TriLens is not to replace $H_x^\ell$ but to supply
complementary pathway-specific readouts under a shared lens.

\paragraph{Superposition and commitment.}
The residual stream at any layer is a linear superposition of all
prior module writes. When the writes are coherent on a single
vocabulary direction, the lens distribution is sharply peaked;
when they are incoherent, the distribution spreads over multiple
candidates.
Shannon entropy is a compact statistic that tracks this spreading: it
is invariant to which candidate tokens are
involved, makes no assumption about whether the vocabulary is flat or
structured, and is a calibrated function of the distribution's
concentration.
Richer alternatives such as KL divergence to the final layer
(DoLa's signal) or a spectral summary lose information about
within-layer commitment, as we show empirically in \S\ref{sec:dola}.

\paragraph{Iterative refinement trajectories.}
The logit-lens view treats each layer as producing a refinement of
a running next-token prediction~\citep{nostalgebraist2020logitlens,belrose2023tunedlens}.
Under this view, $H(\mathrm{lens}(z^\ell))$ is the uncertainty of the
model's belief at depth $\ell$ along that module's pathway.
For a correct continuation, uncertainty often decreases with depth as
module contributions reinforce each other.
For a hallucinated continuation, the refinement trajectory is more
often non-monotone: uncertainty can re-enter the distribution at the
depth where the parametric pathway recalls the true token while the
forced context fixes the output on the wrong one.
The per-layer vector $(H_a^\ell, H_m^\ell, H_x^\ell)_\ell$ can thus be
viewed as a trajectory of this three-pathway refinement process, and a
linear classifier on this trajectory can separate the two regimes in
our benchmark setting.

These three views explain why the feature is minimal (three
location-specific scalars per layer), why it is non-redundant
(each scalar measures a different pathway's commitment), and why it
is architecturally natural (it uses only the model's own unembedding
and pre-trained norm). Section~\ref{sec:exp-perlayer} examines the
resulting empirical layer-wise trajectories.

\subsection{Feature Construction}
\label{sec:method-feature}

Given an evaluation sequence, we run the LLM once and extract
$(H_a^\ell, H_m^\ell, H_x^\ell)$ at the token positions used for
scoring.
Following the standard probe-input convention, we obtain a fixed-length
vector per sample by applying a fixed readout rule consistently
throughout the paper.
The resulting feature is
\begin{equation}
  \mathbf{f} = (H_a^1, H_m^1, H_x^1, \ldots, H_a^L, H_m^L, H_x^L) \in \mathbb{R}^{3L}.
\end{equation}
For the models we consider, $3L$ ranges from 84 (Qwen2.5-7B, $L{=}28$)
to 126 (Gemma-2-9B, $L{=}42$)---roughly $30\times$ smaller than the
raw hidden-state features used by SAPLMA and its descendants.

\subsection{Probes}
\label{sec:method-probe}

We evaluate two classifiers on $\mathbf{f}$:
(i) a \textbf{linear probe} (L2-regularized logistic regression),
which highlights that our feature is already separable without a learned
transformation; and
(ii) an \textbf{MLP probe}
($3L\to128\to64\to32\to1$; LeakyReLU(0.01), BatchNorm, Dropout 0.3),
using the same hidden-layer topology as the architecture used by the
strong prior baseline~\citep{zhang2025icrprobe}. Because TriLens
uses a $3L$-dimensional input whereas ICR Probe uses an $L$-dimensional
input, this comparison controls probe family and training recipe but
does not fully equalize input dimensionality.
We therefore treat the MLP comparison as a matched probe-family
comparison rather than as a dimension-controlled capacity study, and we
also report linear-probe results plus feature-subset ablations to
separate feature quality from probe size. As an additional
dimension-matched control, Appendix~\ref{app:linear} compares TriLens
against a repeated-feature baseline $H_x^{\times 3}$ that simply copies
the residual-stream entropy three times to match the same $3L$ input
size without adding new information, and further reports stricter MLP
controls that reduce the three-pathway feature back to $L$ dimensions
via simple summaries or train-split PCA.
Training details (BCE loss, Adam, 50 epochs, LR scheduler) are
identical to those in~\citet{zhang2025icrprobe} and are listed in
Appendix~\ref{app:hparams}.

\section{Experiments}
\label{sec:exp}

\subsection{Experimental Setup}
\label{sec:exp-setup}

\paragraph{Models.} We evaluate on three recent instruction-tuned LLMs
spanning the $\sim$7--9B parameter range and three architecture families:
Qwen2.5-7B-Instruct~\citep{qwen25}, Meta-Llama-3-8B-Instruct~\citep{llama3},
and Gemma-2-9B-it~\citep{gemma2}.
The first two use the standard RMSNorm+GQA decoder stack;
Gemma-2 alternates full and sliding-window attention.

\paragraph{Datasets.} We evaluate on four benchmarks commonly used in
prior white-box detection work:
HaluEval-QA~\citep{li2023halueval},
SQuAD2.0~\citep{rajpurkar2018squad2},
HotpotQA~\citep{yang2018hotpotqa}, and
TriviaQA~\citep{joshi2017triviaqa}.
For datasets that natively provide positive/negative supervision
(HaluEval, SQuAD2), we use the provided supervision directly.
For HotpotQA and TriviaQA, we follow a fixed benchmark-specific
preprocessing pipeline to instantiate the evaluation examples used in
our experiments. The same preprocessed instances are shared by TriLens
and all baselines.

\paragraph{Metrics and protocol.} Each benchmark is split 80/20 into train
and test sets, and we report test-set AUROC averaged over five random
seeds. Unless otherwise stated, we randomly sample 10,000 instances from
each dataset, keeping the final label distribution balanced within every
dataset. All model states are extracted from a single forward pass over
the evaluation sequence, and the same sampled instances, splits, model
suite, and model--dataset grid are shared by TriLens and all baselines.
Further implementation details are listed in Appendix~\ref{app:hparams}.

\paragraph{Baselines.} We compare against six baselines spanning both
training-free and trainable white-box detectors. The training-free
baselines are perplexity (\textbf{PPL})~\citep{ren2022ood},
length-normalized entropy (\textbf{LN-Entropy})~\citep{malinin2020entropy},
and \textbf{LLM-Check}~\citep{sriramanan2024llmcheck}; the trainable
baselines are \textbf{SAPLMA}~\citep{azaria2023saplma},
\textbf{SEP}~\citep{kossen2024sep}, and the previous state-of-the-art
\textbf{ICR Probe}~\citep{zhang2025icrprobe}. Appendix~\ref{app:baselines}
summarizes these baselines and the evaluation protocol used in our runs.
All baseline results are produced under the same evaluation protocol,
group-preserving 80/20 split, model suite, and model--dataset
grid as TriLens.

\subsection{Main Results}
\label{sec:exp-main}

Table~\ref{tab:main} reports test AUROC on all (model, dataset) cells
against the six baselines.
Our primary method, \textbf{TriLens (MLP)} (last row per model), uses the feature
$\mathbf{f} = (H_a^\ell, H_m^\ell, H_x^\ell)_\ell$ with the same
MLP probe topology used by ICR Probe. Unless otherwise noted,
\emph{TriLens} refers to this MLP variant used in Table~\ref{tab:main}.

\begin{table*}[!t]
  \centering
  \small
  \setlength{\tabcolsep}{8pt}
  \begin{tabular}{@{}c|c|cccc@{}}
    \toprule
    \textbf{LLM} & \textbf{Methods}        & \textbf{HaluEval}  & \textbf{SQuAD2}    & \textbf{HotpotQA}  & \textbf{TriviaQA}  \\
    \midrule
    \multirow{7}{*}{\textbf{Gemma-2-9B}}
                 & PPL                     & 0.5538             & 0.5348             & 0.7239             & 0.7536             \\
                 & LN-Entropy              & 0.7357             & 0.6818             & 0.7349             & 0.7023             \\
                 & LLM-Check               & 0.5780             & 0.5517             & 0.5102             & 0.5911             \\
                 & SAPLMA                  & 0.8123             & 0.7409             & 0.8329             & 0.7766             \\
                 & SEP                     & 0.6439             & 0.6627             & 0.6149             & 0.7726             \\
                 & ICR Probe               & \underline{0.8436} & \underline{0.8142} & \underline{0.8409} & \underline{0.8001} \\
                 & \textbf{TriLens (Ours)} & \textbf{0.9277}    & \textbf{0.9270}    & \textbf{0.9433}    & \textbf{0.9106}    \\
    \midrule
    \multirow{7}{*}{\textbf{Qwen2.5-7B}}
                 & PPL                     & 0.5512             & 0.5278             & 0.6069             & 0.7041             \\
                 & LN-Entropy              & 0.7286             & 0.6564             & 0.6913             & 0.6938             \\
                 & LLM-Check               & 0.5367             & 0.5639             & 0.5518             & 0.5604             \\
                 & SAPLMA                  & 0.7725             & 0.7016             & 0.7689             & \textbf{0.8297}    \\
                 & SEP                     & 0.6634             & 0.6418             & 0.6536             & 0.7449             \\
                 & ICR Probe               & \underline{0.8076} & \underline{0.7382} & \underline{0.7865} & 0.7751             \\
                 & \textbf{TriLens (Ours)} & \textbf{0.9053}    & \textbf{0.9061}    & \textbf{0.9242}    & \underline{0.8103} \\
    \midrule
    \multirow{7}{*}{\textbf{Llama-3-8B}}
                 & PPL                     & 0.5867             & 0.6472             & 0.6658             & 0.7085             \\
                 & LN-Entropy              & 0.6574             & 0.6249             & 0.6661             & 0.5928             \\
                 & LLM-Check               & 0.5263             & 0.5356             & 0.5559             & 0.5482             \\
                 & SAPLMA                  & 0.7315             & 0.7043             & \underline{0.7768} & 0.7586             \\
                 & SEP                     & 0.7309             & \underline{0.7274} & 0.6607             & 0.7048             \\
                 & ICR Probe               & \underline{0.7671} & 0.7568             & 0.7909             & 0.7396             \\
                 & \textbf{TriLens (Ours)} & \textbf{0.9136}    & \textbf{0.9169}    & \textbf{0.9425}    & \textbf{0.8861}    \\
    \bottomrule
  \end{tabular}
  \caption{\textbf{Main results}: test AUROC on four benchmark datasets
    under a matched evaluation protocol. \textbf{Bold} marks the best
    method in each model--dataset cell; \underline{underlining} marks the
    second-best. TriLens improves over ICR Probe on all 12 cells and is
    best in 11, with average gain $+12.1$ AUROC. Per-cell standard
    deviations are listed in Appendix~\ref{app:main-std}.}
  \label{tab:main}
\end{table*}

\begin{figure}[!t]
  \centering
  \includegraphics[width=\columnwidth]{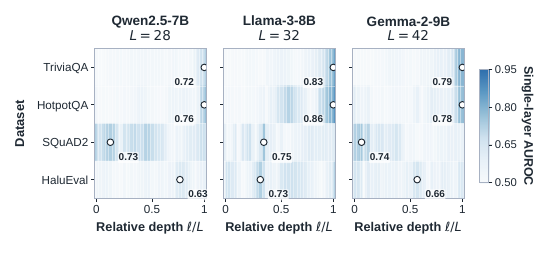}
  \caption{Single-layer AUROC heatmap from $H_x^\ell$. Rows are
    benchmarks, columns are relative depth $\ell/L$, and the peak cell in
    each row is circled. Peak-discriminative depth varies systematically
    across models.}
  \label{fig:heatmap}
\end{figure}

\paragraph{Main finding.}
Hallucination risk is visible in the way internal certainty forms across
layers: a compact trajectory of logit-lens entropies is sufficient to
separate supported from hallucinated answers in a single forward pass.
TriLens improves over ICR Probe on all 12 cells and is the highest-AUROC
entry in 11 of the 12 cells, with gains ranging from $+4.2$ AUROC
(Qwen2.5-7B on TriviaQA) to $+16.0$ AUROC (Qwen2.5-7B on SQuAD2).
The improvement is consistent across model families ($+10.2$ averaged
over Gemma-2, $+11.0$ over Qwen2.5, and $+15.1$ over Llama-3) and across
benchmark regimes ($+12.8$ over the three context-grounded tasks and
$+10.2$ on TriviaQA).

\paragraph{Probe and capacity controls.}
The pattern is not explained solely by the MLP probe. A linear probe on
the same features also outperforms ICR Probe on all 12 cells, with the
MLP adding another $\sim 3.2$ AUROC points (Appendix~\ref{app:linear}).
Dimension-matched controls further show that simply repeating
$H_x$ three times is nearly identical to $H_x$ alone, whereas TriLens
remains better on all 12 linear-probe cells. A stricter MLP sweep with
$L$-dimensional reductions of the three-pathway feature gives the same
conclusion, indicating that the gain is not explained by input
dimensionality alone.

\subsection{Feature Ablation}
\label{sec:exp-ablation}

\begin{figure*}[!t]
  \centering
  \includegraphics[width=0.96\textwidth]{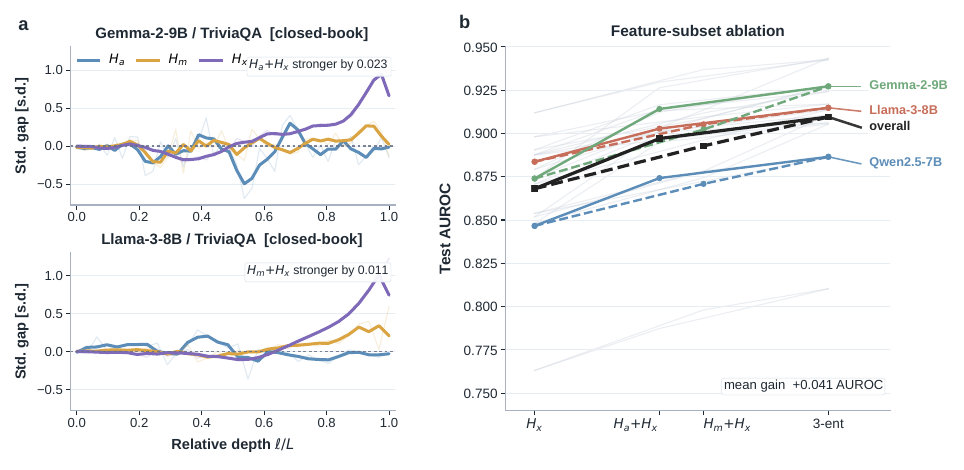}
  \caption{\textbf{Complementarity of the three entropy features.}
    \textbf{Left:} standardized per-layer separation between hallucinated
    and correct samples for $H_a^\ell$, $H_m^\ell$, and $H_x^\ell$ on two
    representative TriviaQA cells. The stronger two-feature branch is
    architecture-dependent. \textbf{Right:} feature-subset ablation across
    all 12 (model, dataset) cells. Colored lines show per-model means; grey
    lines show individual cells. Both two-feature branches improve over
    $H_x$ alone, and the full three-entropy feature attains the best mean.}
  \label{fig:complementarity}
\end{figure*}

The ablation answers whether the three readouts provide redundant or
complementary evidence. We train the MLP probe on progressively larger
subsets of $\mathbf{f}$ and additionally include an intra-layer
module-disagreement term
$\mathrm{JSD}_{am}^\ell = \mathrm{JSD}(\mathrm{lens}(a_i^\ell), \mathrm{lens}(m_i^\ell))$
as a fifth feature, motivated by an intra-layer analog of ICR Probe's
consistency construction.
Detailed ablation numbers are reported in Appendix~\ref{app:ablation}.
The useful signal is not confined to the residual stream: adding either
$H_a$ or $H_m$ to $H_x$ improves performance on all 12 cells, and the
full 3-entropy feature improves further in every cell. This indicates
complementary predictive evidence across the three locations. Adding the
intra-layer disagreement term changes AUROC by at most $+0.006$ per cell
and $+0.001$ on average, suggesting little value beyond the entropy
feature itself. The relative importance of $H_a$ and $H_m$ is
architecture-dependent: the $H_a{+}H_x$ branch is stronger on Gemma-2,
whereas $H_m{+}H_x$ is stronger on Llama-3 and part of Qwen2.5.

Figure~\ref{fig:complementarity} visualizes this complementarity from
two angles. The left panel shows that the three entropies open
hallucination--correct separation at different depths and with
architecture-specific pathway asymmetry; the right panel shows that
both two-feature branches improve over $H_x$ alone, while the full
3-entropy feature attains the highest overall mean.

\subsection{Per-Layer Analysis}
\label{sec:exp-perlayer}
Detailed trajectory plots and peak-layer densities are deferred to
Appendix~\ref{app:perlayer}. In the main text, we focus on the summary
heatmap in Figure~\ref{fig:heatmap}. It shows that the
peak-discriminative depth varies substantially across model--dataset
pairs: some open-book cells peak in early or middle layers, while
TriviaQA and several HotpotQA cells peak only near the final layer.
This benchmark-conditional variation is consistent with prior
mechanistic analyses~\citep{geva2023dissecting,zhang2025icrprobe},
which found that different models integrate knowledge at different
depths.

\subsection{Cross-Dataset Diagnostics}
\label{sec:exp-crossdata}

The main results in Table~\ref{tab:main} use a separate probe per
benchmark. We next use two benchmark-shift diagnostics; detailed results
appear in Appendix~\ref{app:crossdata}. First, we train one probe per
model on the union of all four benchmark training splits and evaluate it
on each benchmark's held-out test split. This shared-probe setting still
exceeds per-dataset ICR Probe on all 12 cells, with average gain $+10.7$
AUROC, while trailing our own per-dataset probes by only $\sim 1.4$ AUROC
on average. Second, we train on one benchmark and test on another to form
benchmark-level transfer heatmaps. In this cross-dataset transfer
diagnostic, TriLens is the strongest method on all 12 off-diagonal cells
and on all 16 cells overall; its off-diagonal mean AUROC is $0.848$,
compared with $0.738$ for ICR Probe and $0.678$ for SAPLMA. We therefore
view the transfer result as further evidence that the gain is not confined
to in-domain fitting, while still treating it as a distribution-shift
diagnostic rather than as evidence of universal cross-benchmark
generalization.

\begin{figure*}[!t]
  \centering
  \includegraphics[width=0.94\textwidth]{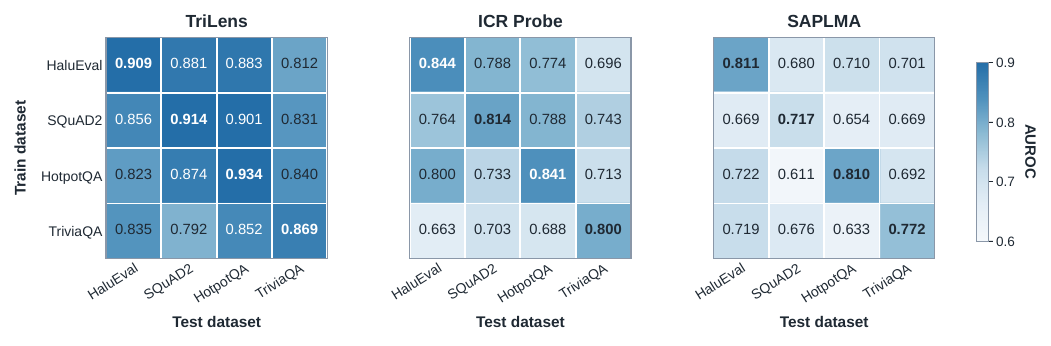}
  \caption{Cross-dataset generalization heatmaps for TriLens, ICR Probe,
    and SAPLMA. Each cell reports AUROC for the train-on-row,
    test-on-column setting. TriLens is strongest on all 12 off-diagonal
    cells and on all 16 cells overall, indicating more stable transfer
    under benchmark shift.}
  \label{fig:crossdata}
\end{figure*}

\subsection{Comparison with Cross-Layer Contrast Signals}
\label{sec:dola}

TriLens and DoLa~\citep{chuang2024dola} both derive signals from
logit-lens distributions but differ in what they measure: DoLa
contrasts logit-lens distributions \emph{across} layers and uses the
result to modify decoding, while we measure entropy \emph{within} each
layer and use it for detection.
To assess whether our per-layer entropy feature could be explained
as a reformulation of the cross-layer contrast signal, we repurpose
DoLa's quantity as a detection feature and compare it head to head
against ours.

Concretely, define the DoLa-style detection feature as
\begin{equation}
  \begin{aligned}
    f_{\mathrm{DoLa}}^\ell
     & = \mathrm{JSD}\!\big(\mathrm{lens}(x^\ell), \mathrm{lens}(x^L)\big), \\
     & \qquad \ell = 0, \ldots, L-1.
  \end{aligned}
\end{equation}
and assemble the $L$-dimensional vector
$\mathbf{f}_{\mathrm{DoLa}}$. We train the same MLP probe on
(a) $\mathbf{f}_{\mathrm{DoLa}}$, (b) the TriLens feature
$\mathbf{f}_{3\text{-ent}}$, and (c) their concatenation
$\mathbf{f}_{3\text{-ent}} \oplus \mathbf{f}_{\mathrm{DoLa}}$.
Experiments are run under the same group-preserving 80/20 evaluation
protocol used in our main experiments, using the same 10k-scale feature
files as the main paper. Table~\ref{tab:dola} reports the main-setting
comparison, while Appendix~\ref{app:dola} gives the full configuration
grid.

\begin{table}[t]
  \centering
  \small
  \setlength{\tabcolsep}{5pt}
  \begin{tabular}{@{}l|ccc@{}}
    \toprule
    \textbf{Dataset} & \textbf{DoLa-JSD} & \textbf{TriLens} & \textbf{TriLens+DoLa} \\
    \midrule
    HaluEval         & 0.7623            & 0.9136           & \textbf{0.9155}       \\
    SQuAD2           & 0.8538            & 0.9158           & \textbf{0.9172}       \\
    HotpotQA         & 0.8928            & \textbf{0.9425}  & 0.9422                \\
    TriviaQA         & 0.8679            & \textbf{0.8861}  & 0.8860                \\
    \bottomrule
  \end{tabular}
  \caption{Head-to-head comparison against DoLa-style cross-layer
    contrast features (MLP probe, test AUROC). Full results appear in
    Appendix~\ref{app:dola}.}
  \label{tab:dola}
\end{table}

\paragraph{Findings.}
Table~\ref{tab:dola} presents three observations.
First, \textbf{our per-layer entropy consistently outperforms the DoLa-style
  cross-layer contrast feature on every dataset}, with gaps ranging from
$+0.02$ to $+0.15$ AUROC and an average gain of roughly $+0.07$ in the
main setting.
Second, \textbf{their union changes performance only slightly and without a
  consistent upward pattern}
relative to our feature alone: averaged across the four datasets, the
union changes AUROC by less than $+0.001$. The cross-layer contrast
signal therefore contributes little beyond what is already captured by
per-layer entropy.
Third, the DoLa-style feature degrades most severely on HaluEval, where
in our default setup, intermediate layers have already converged
onto the final layer's prediction (so the JSD collapses toward zero
regardless of whether the response is hallucinated or correct);
per-layer entropy is not subject to this degeneracy because it
measures the shape of each layer's distribution independently.

\section{Conclusion}

We propose \emph{TriLens}, a $3L$-dimensional
feature that computes Shannon entropies of logit-lens distributions at
the MHSA output, FFN output, and residual stream at every layer.
Across 3 LLMs $\times$ 4 benchmarks, under the same MLP probe
family used by a strong prior baseline, this feature
improves over ICR Probe on all 12 cells with an average gain of
$+12.1$ AUROC and attains the highest AUROC in 11 of the 12 cells.
Ablation identifies all three module entropies as independently
informative and rules out intra-layer module-disagreement as redundant.
A head-to-head comparison suggests that our feature captures much of the
useful signal in DoLa-style cross-layer contrast.
Layer-wise analysis further indicates that the most informative depth
varies across architectures and benchmarks rather than concentrating at
a universal layer. Taken together, these findings suggest that simple,
layer-wise features merit consideration alongside more
elaborate probe architectures in white-box hallucination detection.

\section*{Limitations}

TriLens provides an effective white-box detection signal, but it has
several limitations. First, it requires access to internal activations
and is therefore restricted to open-source or otherwise transparent
models. Second, our study focuses on detection rather than mitigation,
so it does not directly address how such signals should be used to
reduce hallucinations during generation. Finally, although our
experiments span multiple benchmarks and model families, extending the
evaluation to broader model scales and more complex generation settings
would provide a stronger test of robustness and generality.

\section*{Ethics Statement}

Our study is conducted entirely on publicly available LLMs and established
public benchmarks, introduces no new human-subject data collection, and
requires no additional annotation. We do not release any new text dataset
containing personally identifying information or offensive content; all
reported results are aggregate metrics over benchmark splits.
We use the public models and benchmarks only under their publicly stated
access conditions and licenses or terms of use, and our released code
contains no redistributed model weights or benchmark text. Our use of
these artifacts is limited to research evaluation of hallucination
detection and is consistent with their benchmark or model-evaluation
purpose and access conditions.
We acknowledge the standard dual-use consideration that more accurate
hallucination detection could be used to identify, and thereby
refine, adversarial generations. However, as a detection-only
capability without any generation or mitigation component, the net
effect is defensive.

\bibliography{custom}

\appendix

\section{Experimental Details}
\label{app:exp-details}

\subsection{Computing Infrastructure}
\label{app:compute}

Our experiments were conducted on a server equipped with 6 NVIDIA
GeForce RTX 4090 GPUs (24 GB memory each), running PyTorch with CUDA
acceleration. Across multiple rounds of feature extraction, ablation,
and evaluation experiments, the total computational budget amounted to
100--200 GPU hours. Feature extraction requires one teacher-forced
forward pass per scored sequence; probe training is lightweight relative
to LLM-state extraction and is repeated across seeds and evaluation
configurations reported in the main text and appendix.

\subsection{Details about Datasets}
\label{app:datasets}

For our experiments, we randomly sampled 10,000 instances from each
dataset. Specifically, we used the QA subset of HaluEval and the
`rc.nocontext' subset of TriviaQA. These benchmarks are English-language
QA-style evaluation sets covering factoid question answering,
unanswerable questions, multi-hop reasoning, and hallucinated QA
responses; we do not use demographic annotations or perform demographic
subgroup analysis. Each dataset is split into 80\%-20\% for training and
testing. We train and test on each dataset, reporting the corresponding
results. This experimental setup for baseline methods matches ours
exactly.

\subsection{Sampling, Splits, and Supervision}
\label{app:splits}

All trainable methods use group-preserving 80/20 train/test splits, and
the same split assignment is shared by TriLens and all baselines. All
reported results are averaged over five random seeds.

\subsection{Details about Baselines}
\label{app:baselines}

We summarize the six baselines in Table~\ref{tab:main}. The first three
are training-free sample-level scorers, whereas the latter three are
trainable white-box detectors. All are evaluated in the same pipeline as
TriLens, on the same model--dataset grid and group-preserving 80/20
split. Scalar baselines are used directly as sample-level
scores; trainable baselines retain their original feature definition and
are fit on the corresponding training split.
For the 10k-scale evaluation sets used in the main paper, HaluEval and SQuAD2
follow the dataset sizes reported in the main text.

\paragraph{PPL.}
Perplexity of the scored sequence under the same forward-pass setup.
Higher
perplexity indicates lower model confidence and is used directly as a
hallucination score.

\paragraph{LN-Entropy.}
Length-normalized predictive entropy of the scored sequence. This baseline
measures output uncertainty while partially controlling for raw response
length.

\paragraph{LLM-Check.}
A training-free attention-kernel baseline that scores responses from
structural statistics of self-attention kernels rather than hidden
states.

\paragraph{SAPLMA.}
A supervised hidden-state baseline that trains an MLP classifier on
internal activations from selected layers.

\paragraph{SEP.}
A probe-based hidden-state baseline motivated by semantic entropy,
using learned detectors over internal representations rather than
black-box semantic clustering over sampled outputs.

\paragraph{ICR Probe.}
The strongest prior white-box baseline. It derives one scalar feature
per layer from residual-stream update statistics and feeds the resulting
layer-wise vector to the same 4-layer MLP architecture used throughout
our comparisons; the shared probe hyperparameters are listed in
Appendix~\ref{app:hparams}.

\subsection{Probe Comparability Note}
\label{app:probe-compare}

TriLens and ICR Probe are compared under the same training protocol and
the same hidden-layer MLP topology, but not under equal input
dimensionality: TriLens uses $3L$ input features whereas ICR Probe uses
$L$. The main text therefore interprets the comparison as a
feature-plus-probe comparison within a matched probe family, rather than
as a fully dimension-controlled capacity study.

\section{Probe Controls and Fairness Checks}
\label{app:linear}

Table~\ref{tab:linear} reports the same TriLens feature used in the
main paper, replacing the MLP probe with L2-regularized logistic
regression under the same feature configuration as the main results.
The qualitative conclusion is unchanged: the linear probe already
improves over ICR Probe on all 12 cells, with the MLP adding a further
$\sim 3.2$ AUROC points on average. To isolate the effect of input
dimensionality, Table~\ref{tab:linear-fairness} adds a
dimension-matched control $H_x^{\times 3}$ that repeats the
residual-stream entropy feature three times, yielding the same $3L$
input size as TriLens without introducing any new signal.
Table~\ref{tab:mlp-fairness} extends this check under the same MLP
probe family with stricter dimensionality controls: three handcrafted
$L$-dimensional reductions of the three-pathway feature and a train-split
PCA projection of the full $3L$-dimensional TriLens vector back to $L$.
We further report alternative readout controls in
Appendix~\ref{app:readout}, comparing the pathway-separated TriLens
readout against intermediate-state and additive-write readouts built
from the same layer-wise logit-lens pipeline.

\begin{table*}[t]
  \centering
  \small
  \setlength{\tabcolsep}{8pt}
  \begin{tabular}{@{}c|c|cccc@{}}
    \toprule
    \textbf{LLM} & \textbf{Methods}          & \textbf{HaluEval} & \textbf{SQuAD2} & \textbf{HotpotQA} & \textbf{TriviaQA} \\
    \midrule
    \multirow{2}{*}{\textbf{Gemma-2-9B}}
                 & ICR Probe                 & 0.8436            & 0.8142          & 0.8409            & 0.8001            \\
                 & \textbf{TriLens (Linear)} & \textbf{0.8963}   & \textbf{0.9004} & \textbf{0.9203}   & \textbf{0.8864}   \\
    \midrule
    \multirow{2}{*}{\textbf{Qwen2.5-7B}}
                 & ICR Probe                 & 0.8003            & 0.7456          & 0.7917            & 0.7684            \\
                 & \textbf{TriLens (Linear)} & \textbf{0.8597}   & \textbf{0.8696} & \textbf{0.8746}   & \textbf{0.7780}   \\
    \midrule
    \multirow{2}{*}{\textbf{Llama-3-8B}}
                 & ICR Probe                 & 0.7603            & 0.7634          & 0.7982            & 0.7325            \\
                 & \textbf{TriLens (Linear)} & \textbf{0.8731}   & \textbf{0.8806} & \textbf{0.9253}   & \textbf{0.8700}   \\
    \bottomrule
  \end{tabular}
  \caption{Linear-probe results on the same 12 cells as
    Table~\ref{tab:main}. All TriLens entries are 5-seed means; seed
    standard deviation is $\leq 0.007$ in every cell.}
  \label{tab:linear}
\end{table*}

\begin{table}[t]
  \centering
  \small
  \setlength{\tabcolsep}{5pt}
  \resizebox{\columnwidth}{!}{%
    \begin{tabular}{@{}c|c|ccc@{}}
      \toprule
      \textbf{LLM} & \textbf{Dataset} & $H_x$  & $H_x^{\times 3}$ & \textbf{TriLens} \\
      \midrule
      \multirow{4}{*}{\textbf{Gemma-2-9B}}
                   & HaluEval         & 0.7970 & 0.7970           & \textbf{0.8963}  \\
                   & SQuAD2           & 0.8497 & 0.8497           & \textbf{0.9004}  \\
                   & HotpotQA         & 0.7812 & 0.7813           & \textbf{0.9203}  \\
                   & TriviaQA         & 0.8238 & 0.8238           & \textbf{0.8864}  \\
      \midrule
      \multirow{4}{*}{\textbf{Qwen2.5-7B}}
                   & HaluEval         & 0.7746 & 0.7746           & \textbf{0.8597}  \\
                   & SQuAD2           & 0.8389 & 0.8389           & \textbf{0.8696}  \\
                   & HotpotQA         & 0.7992 & 0.7992           & \textbf{0.8746}  \\
                   & TriviaQA         & 0.7384 & 0.7384           & \textbf{0.7780}  \\
      \midrule
      \multirow{4}{*}{\textbf{Llama-3-8B}}
                   & HaluEval         & 0.8247 & 0.8248           & \textbf{0.8731}  \\
                   & SQuAD2           & 0.8254 & 0.8255           & \textbf{0.8806}  \\
                   & HotpotQA         & 0.8717 & 0.8717           & \textbf{0.9253}  \\
                   & TriviaQA         & 0.8385 & 0.8385           & \textbf{0.8700}  \\
      \bottomrule
    \end{tabular}
  }
  \caption{Dimension-matched fairness control under the linear probe.
  $H_x^{\times 3}$ repeats the $L$-dimensional residual-stream entropy
  feature three times to match TriLens's $3L$ input size. The repeated
  baseline is nearly identical to $H_x$, while TriLens remains better on
  all 12 cells.}
  \label{tab:linear-fairness}
\end{table}

\begin{table*}[t]
  \centering
  \small
  \setlength{\tabcolsep}{5pt}
  \resizebox{\textwidth}{!}{%
    \begin{tabular}{@{}c|c|ccccccc@{}}
      \toprule
      \textbf{LLM} & \textbf{Dataset} & $H_x$  & $H_x^{\times 3}$ & \textbf{TriMeanL} & \textbf{TriMaxL} & \textbf{TriMinL} & \textbf{TriLens+PCA$\to L$} & \textbf{TriLens} \\
      \midrule
      \multirow{4}{*}{\textbf{Gemma-2-9B}}
                   & HaluEval         & 0.8325 & 0.8420           & 0.8364            & 0.8140           & 0.8188           & 0.8794                      & \textbf{0.8967}  \\
                   & SQuAD2           & 0.8532 & 0.8638           & 0.8461            & 0.8319           & 0.8349           & 0.8773                      & \textbf{0.8951}  \\
                   & HotpotQA         & 0.8051 & 0.8272           & 0.8326            & 0.7991           & 0.8050           & 0.8927                      & \textbf{0.9145}  \\
                   & TriviaQA         & 0.8237 & 0.8313           & 0.8057            & 0.7681           & 0.8053           & 0.8714                      & \textbf{0.8838}  \\
      \midrule
      \multirow{4}{*}{\textbf{Qwen2.5-7B}}
                   & HaluEval         & 0.8000 & 0.8131           & 0.8004            & 0.7534           & 0.7877           & 0.8379                      & \textbf{0.8576}  \\
                   & SQuAD2           & 0.8523 & 0.8606           & 0.8358            & 0.8134           & 0.8202           & 0.8669                      & \textbf{0.8743}  \\
                   & HotpotQA         & 0.8376 & 0.8529           & 0.8155            & 0.7813           & 0.7585           & 0.8655                      & \textbf{0.8789}  \\
                   & TriviaQA         & 0.7330 & 0.7387           & 0.6933            & 0.6218           & 0.6934           & 0.7667                      & \textbf{0.7700}  \\
      \midrule
      \multirow{4}{*}{\textbf{Llama-3-8B}}
                   & HaluEval         & 0.7635 & 0.7783           & 0.7874            & 0.7480           & 0.7840           & 0.8209                      & \textbf{0.8379}  \\
                   & SQuAD2           & 0.8100 & 0.8383           & 0.7902            & 0.7469           & 0.8077           & 0.8093                      & \textbf{0.8448}  \\
                   & HotpotQA         & 0.8612 & 0.8710           & 0.8737            & 0.8057           & 0.8539           & 0.9034                      & \textbf{0.9156}  \\
                   & TriviaQA         & 0.8305 & 0.8326           & 0.8130            & 0.6753           & 0.8110           & 0.8471                      & \textbf{0.8578}  \\
      \bottomrule
    \end{tabular}
  }
  \caption{Stricter dimensionality controls under the MLP probe.
  Besides $H_x^{\times 3}$, we compare three handcrafted
  $L$-dimensional reductions of the three-pathway feature and a
  train-split PCA projection of the full $3L$-dimensional TriLens vector
  back to $L$. Full TriLens remains strongest on all 12 cells.}
  \label{tab:mlp-fairness}
\end{table*}

\subsection{Alternative Readout Controls}
\label{app:readout}

The main TriLens feature applies the final logit-lens readout to the
isolated MHSA write $a^\ell$, the isolated FFN write $m^\ell$, and the
composed residual state $x^\ell$. To test whether the gains depend on
this particular choice of readout location, we compare it with two more
conservative alternatives built from the same layer-wise pipeline:
(i) $H_{\mathrm{pre}}$, the entropy of the intermediate state
$x^{\ell-1}+a^\ell$ after the attention write but before the FFN write,
and (ii) $H_p$, the entropy of the additive-write state
$a^\ell + m^\ell$. We then pair each alternative with the standard
residual-stream readout $H_x$ and evaluate the resulting features under
the same protocol as the main paper.

\begin{table}[t]
  \centering
  \small
  \setlength{\tabcolsep}{4pt}
  \begin{tabular}{@{}lcc|cc@{}}
    \toprule
    \multirow{2}{*}{\textbf{Readout}} & \multicolumn{2}{c|}{\textbf{Linear}} & \multicolumn{2}{c}{\textbf{MLP}}                                    \\
                                      & \textbf{Avg}                         & \textbf{Wins}                    & \textbf{Avg}    & \textbf{Wins}  \\
    \midrule
    $H_x$                             & 0.8137                               & 0/12                             & 0.8169          & 0/12           \\
    $H_{\mathrm{pre}}+H_x$            & 0.8334                               & 0/12                             & 0.8268          & 0/12           \\
    $H_p+H_x$                         & 0.8604                               & 0/12                             & 0.8496          & 1/12           \\
    \textbf{TriLens}                  & \textbf{0.8779}                      & \textbf{12/12}                   & \textbf{0.8690} & \textbf{11/12} \\
    \bottomrule
  \end{tabular}
  \caption{Readout-location robustness summary across the 12
    model--dataset cells. ``Avg'' denotes mean test AUROC; ``Wins'' counts
    how often a readout is the strongest cell-wise configuration. All
    entries are 5-seed means under the same protocol as the main paper.}
  \label{tab:readout}
\end{table}

Three patterns are consistent across probe families. First,
performance improves monotonically as the readout moves from
residual-only $H_x$ to $H_{\mathrm{pre}}+H_x$, then to $H_p+H_x$, and
finally to the full pathway-separated TriLens feature. Under the linear
probe, the corresponding average AUROC increases from $0.814$ to
$0.833$, $0.860$, and $0.878$, with TriLens strongest on all 12 cells.
Second, the same ordering largely persists under the MLP probe, where
TriLens is strongest on 11 of 12 cells and improves over $H_x$ by
$+0.052$ AUROC on average. Third, the mixed-state alternatives do help,
but they remain consistently below the explicit three-way decomposition.
We therefore interpret this sweep as evidence that TriLens benefits from
separating MHSA, FFN, and residual-stream readouts, rather than from
using an arbitrary auxiliary lens location.

\section{Cross-Dataset Generalization Heatmaps}
\label{app:crossdata}

The main paper briefly reports the shared-probe variant of TriLens.
Table~\ref{tab:multi-dataset} gives the corresponding per-benchmark
numbers. We additionally report benchmark-level cross-dataset
generalization heatmaps for TriLens, ICR Probe, and SAPLMA. In each
$4{\times}4$ matrix, a probe is trained on the row benchmark and
evaluated on the column benchmark under the same train/test transfer
protocol.

\begin{table}[t]
  \centering
  \small
  \setlength{\tabcolsep}{4pt}
  \resizebox{\columnwidth}{!}{%
    \begin{tabular}{@{}l|l|cc|c@{}}
      \toprule
      \textbf{Model} & \textbf{Test} & \textbf{ICR}\textsuperscript{\dag} & \textbf{Multi (ours)} & $\Delta$          \\
      \midrule
      \multirow{4}{*}{Qwen2.5-7B}
                     & HaluEval      & 0.8003                             & 0.8856                & $+0.085$          \\
                     & SQuAD2        & 0.7456                             & 0.8915                & $+0.146$          \\
                     & HotpotQA      & 0.7917                             & 0.9102                & $+0.118$          \\
                     & TriviaQA      & 0.7684                             & 0.7998                & $+0.031$          \\
      \midrule
      \multirow{4}{*}{Llama-3-8B}
                     & HaluEval      & 0.7603                             & 0.8965                & $+0.136$          \\
                     & SQuAD2        & 0.7634                             & 0.9045                & $+0.141$          \\
                     & HotpotQA      & 0.7982                             & 0.9329                & $+0.135$          \\
                     & TriviaQA      & 0.7325                             & 0.8739                & $+0.141$          \\
      \midrule
      \multirow{4}{*}{Gemma-2-9B}
                     & HaluEval      & 0.8436                             & 0.9132                & $+0.070$          \\
                     & SQuAD2        & 0.8142                             & 0.9129                & $+0.099$          \\
                     & HotpotQA      & 0.8409                             & 0.9225                & $+0.082$          \\
                     & TriviaQA      & 0.8001                             & 0.8971                & $+0.097$          \\
      \midrule
      \textbf{avg}   &               & 0.7883                             & \textbf{0.8950}       & $\mathbf{+0.107}$ \\
      \bottomrule
    \end{tabular}
  }
  \caption{\textbf{Multi-dataset probe}: one probe per model, trained on
    the union of all four benchmarks and evaluated on each held-out test
    split. Each cell is a 5-seed mean (std $\leq 0.008$).
    \textsuperscript{\dag}ICR entries are our per-dataset reruns from
    Table~\ref{tab:main}.}
  \label{tab:multi-dataset}
\end{table}

\paragraph{Aggregate statistics.}
Across the 12 off-diagonal cells, TriLens reaches a mean AUROC of
$0.848$, compared with $0.738$ for ICR Probe and $0.678$ for SAPLMA.
Its diagonal mean is also higher ($0.907$ vs.\ $0.825$ and $0.778$),
so the advantage is not limited to either in-domain or out-of-domain
evaluation alone.

\paragraph{Transfer pattern.}
The transfer gap is not perfectly uniform across source benchmarks.
For TriLens, probes trained on SQuAD2 or HaluEval give the strongest
off-diagonal averages ($0.863$ and $0.859$), while TriviaQA is the
weakest source row ($0.826$ off-diagonal mean). Even so, all 12
off-diagonal TriLens cells remain above $0.79$, whereas the baselines
show broader degradation. We therefore interpret Figure~\ref{fig:crossdata}
as evidence of comparatively stable transfer under benchmark shift,
while still viewing cross-dataset evaluation as a diagnostic rather
than as a substitute for the in-domain results in Table~\ref{tab:main}.

\section{Main-Table Standard Deviations}
\label{app:main-std}

Table~\ref{tab:main-std} lists the standard deviations for the TriLens
entries in Table~\ref{tab:main}, across the same five seeds used in the
main-paper evaluation.

\begin{table}[t]
  \centering
  \small
  \setlength{\tabcolsep}{4pt}
  \begin{tabular}{@{}llc@{}}
    \toprule
    \textbf{Model} & \textbf{Dataset} & \textbf{Std} \\
    \midrule
    Gemma-2-9B     & HaluEval         & 0.0022       \\
    Gemma-2-9B     & SQuAD2           & 0.0025       \\
    Gemma-2-9B     & HotpotQA         & 0.0032       \\
    Gemma-2-9B     & TriviaQA         & 0.0027       \\
    \midrule
    Qwen2.5-7B     & HaluEval         & 0.0053       \\
    Qwen2.5-7B     & SQuAD2           & 0.0025       \\
    Qwen2.5-7B     & HotpotQA         & 0.0015       \\
    Qwen2.5-7B     & TriviaQA         & 0.0051       \\
    \midrule
    Llama-3-8B     & HaluEval         & 0.0014       \\
    Llama-3-8B     & SQuAD2           & 0.0037       \\
    Llama-3-8B     & HotpotQA         & 0.0033       \\
    Llama-3-8B     & TriviaQA         & 0.0037       \\
    \bottomrule
  \end{tabular}
  \caption{Seed standard deviations for the TriLens entries in
    Table~\ref{tab:main}.}
  \label{tab:main-std}
\end{table}

\section{Feature Ablation}
\label{app:ablation}

Table~\ref{tab:ablation} reports the full feature-set ablation from the
main paper discussion.

\begin{table}[t]
  \centering
  \small
  \setlength{\tabcolsep}{3pt}
  \resizebox{\columnwidth}{!}{%
    \begin{tabular}{@{}ll|ccccc@{}}
      \toprule
      \textbf{Model} & \textbf{Data} & $H_x$ & $H_a{+}H_x$ & $H_m{+}H_x$ & \textbf{3-ent} & \textbf{3-ent}$+\mathrm{JSD}_{am}^{\ell}$ \\
      \midrule
      \multirow{4}{*}{Gemma-2}
                     & HaluEval      & .8793 & .9123       & .9087       & .9277          & \textbf{.9286}                            \\
                     & SQuAD2        & .8905 & .9165       & .9085       & .9270          & \textbf{.9285}                            \\
                     & HotpotQA      & .8741 & .9264       & .9135       & .9433          & \textbf{.9451}                            \\
                     & TriviaQA      & .8518 & .9013       & .8782       & .9106          & \textbf{.9115}                            \\
      \midrule
      \multirow{4}{*}{Qwen2.5}
                     & HaluEval      & .8480 & .8950       & .8743       & .9053          & \textbf{.9067}                            \\
                     & SQuAD2        & .8876 & .9006       & .8992       & .9061          & \textbf{.9065}                            \\
                     & HotpotQA      & .8877 & .9138       & .9118       & .9242          & \textbf{.9256}                            \\
                     & TriviaQA      & .7632 & .7874       & .7981       & \textbf{.8103} & .8102                                     \\
      \midrule
      \multirow{4}{*}{Llama-3}
                     & HaluEval      & .8708 & .9008       & .8962       & .9136          & \textbf{.9194}                            \\
                     & SQuAD2        & .8981 & .9132       & .9095       & .9169          & \textbf{.9194}                            \\
                     & HotpotQA      & .9119 & .9296       & .9368       & .9425          & \textbf{.9444}                            \\
                     & TriviaQA      & .8538 & .8675       & .8783       & \textbf{.8861} & .8846                                     \\
      \bottomrule
    \end{tabular}
  }
  \caption{Feature-set ablation (MLP probe, test AUROC).
    \textbf{3-ent} $=(H_a, H_m, H_x)$; the last column adds
    $\mathrm{JSD}_{am}^\ell$.}
  \label{tab:ablation}
\end{table}

Table~\ref{tab:ablation-std} lists the standard deviations across the
same five seeds used in Table~\ref{tab:ablation}.

\begin{table}[t]
  \centering
  \small
  \setlength{\tabcolsep}{3pt}
  \resizebox{\columnwidth}{!}{%
    \begin{tabular}{@{}ll|ccccc@{}}
      \toprule
      \textbf{Model} & \textbf{Data} & $H_x$ & $H_a{+}H_x$ & $H_m{+}H_x$ & \textbf{3-ent} & \textbf{3-ent}$+\mathrm{JSD}_{am}^{\ell}$ \\
      \midrule
      \multirow{4}{*}{Gemma-2}
                     & HaluEval      & 0.004 & 0.002       & 0.003       & 0.002          & 0.002                                     \\
                     & SQuAD2        & 0.006 & 0.004       & 0.006       & 0.004          & 0.004                                     \\
                     & HotpotQA      & 0.003 & 0.003       & 0.003       & 0.004          & 0.004                                     \\
                     & TriviaQA      & 0.004 & 0.004       & 0.004       & 0.004          & 0.005                                     \\
      \midrule
      \multirow{4}{*}{Qwen2.5}
                     & HaluEval      & 0.005 & 0.002       & 0.005       & 0.003          & 0.003                                     \\
                     & SQuAD2        & 0.006 & 0.005       & 0.004       & 0.006          & 0.005                                     \\
                     & HotpotQA      & 0.005 & 0.005       & 0.003       & 0.004          & 0.005                                     \\
                     & TriviaQA      & 0.006 & 0.004       & 0.008       & 0.006          & 0.006                                     \\
      \midrule
      \multirow{4}{*}{Llama-3}
                     & HaluEval      & 0.003 & 0.003       & 0.001       & 0.002          & 0.002                                     \\
                     & SQuAD2        & 0.003 & 0.004       & 0.004       & 0.004          & 0.004                                     \\
                     & HotpotQA      & 0.004 & 0.003       & 0.004       & 0.002          & 0.003                                     \\
                     & TriviaQA      & 0.003 & 0.003       & 0.005       & 0.004          & 0.004                                     \\
      \bottomrule
    \end{tabular}
  }
  \caption{Standard deviations corresponding to Table~\ref{tab:ablation}
    (MLP probe, same aggregation as the main results).}
  \label{tab:ablation-std}
\end{table}
\section{Per-Layer Diagnostics and Mechanistic Reading}
\label{app:perlayer}

This appendix collects the layer-wise plots and the fuller
interpretation omitted from the compressed main text.

Peak-layer analysis is included only as a descriptive diagnostic rather
than as a tuned component of TriLens. The main results always use the
full per-layer feature vector, without selecting or reweighting layers
based on validation performance. We report the peak-discriminative layer
only to summarize where the single-feature $H_x^\ell$ detector reaches
its highest AUROC in each model--dataset cell.

\begin{figure}[t]
  \centering
  \includegraphics[width=\columnwidth]{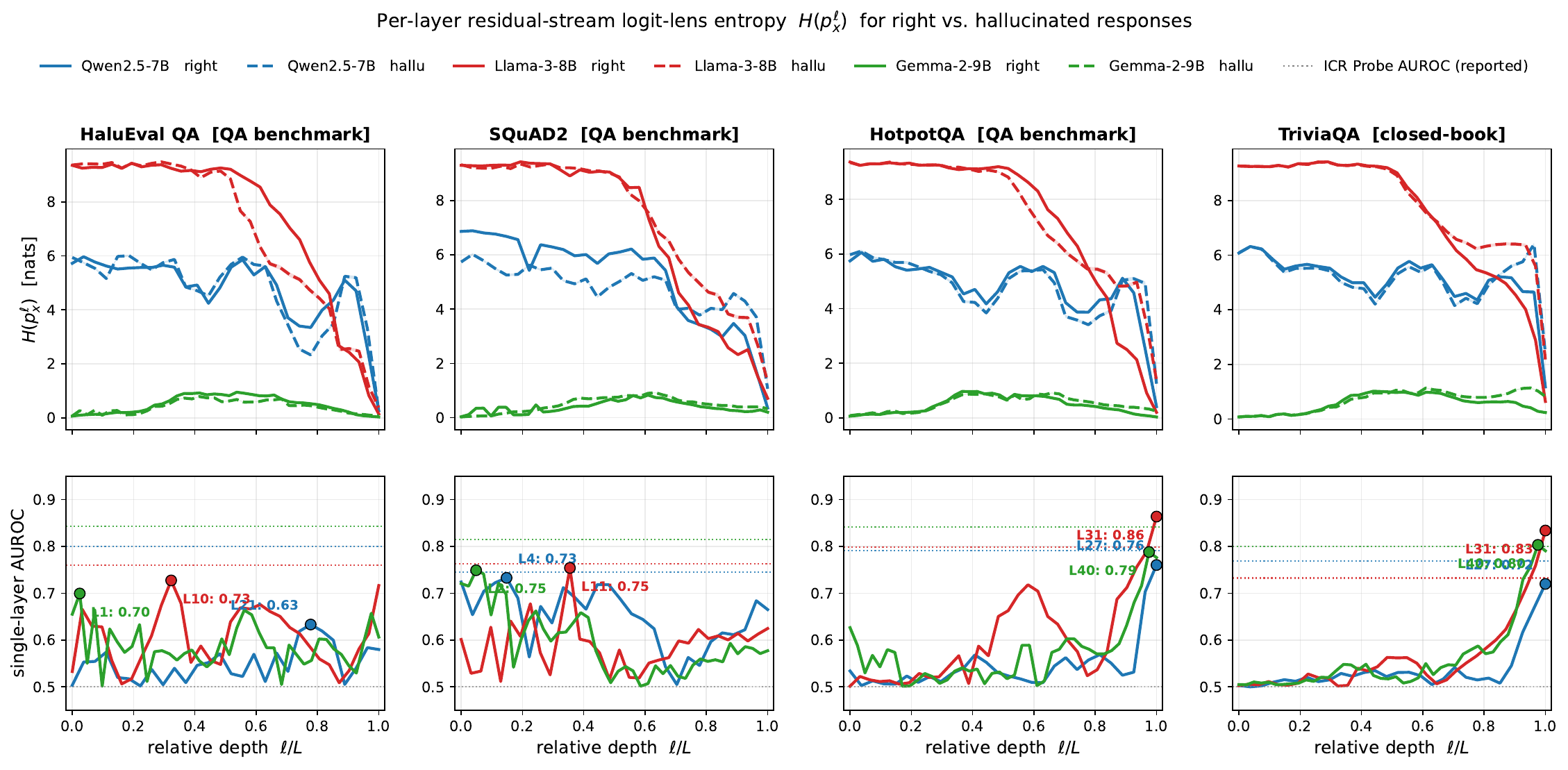}
  \caption{\textbf{Per-layer residual-stream logit-lens entropy
      $H_x^\ell$ (top)} and \textbf{single-layer AUROC (bottom)} across three
    models and four benchmarks. Solid lines denote correct responses;
    dashed lines denote hallucinated responses.}
  \label{fig:mechanism}
\end{figure}

\paragraph{Context-grounded vs.\ closed-book regimes.}
On context-grounded benchmarks such as SQuAD2 and HotpotQA, the
correct--hallucinated gap often opens earlier and remains visible across
much of the upper stack. On TriviaQA, by contrast, the trajectories tend
to remain close until the final layers, consistent with a closed-book
setting in which the relevant parametric recall signal is resolved late.

\paragraph{Architecture-dependent peak depth.}
Figure~\ref{fig:heatmap} shows that the peak-discriminative layer is not
universal across architectures. Some Qwen2.5 cells peak in earlier or
mid layers, whereas Llama-3 and Gemma-2 frequently peak closer to the
final layers. This pattern is consistent with prior observations that
different model families accumulate factual recall at different depths.

\paragraph{Pathway asymmetry.}
The reversal between Gemma-2 and Llama-3 in Figure~\ref{fig:complementarity}
suggests that the relative usefulness of $H_a$ and $H_m$ depends on how
readable each pathway's contribution is under the model's own
unembedding. Gemma-2's hybrid attention stack may shift that balance
toward MHSA relative to the other two models.

\paragraph{Smallest-gain cell.}
The smallest gain in the main table occurs on Qwen2.5-7B / TriviaQA.
Its late-layer trajectories show weaker separation than the
corresponding cells for Llama-3 and Gemma-2, suggesting a more diffuse
parametric recall signal in this closed-book setting.

Table~\ref{tab:peak} reports the peak-discriminative layer for the
single-feature detector based on $H_x^\ell$, using the same setup as
Figures~\ref{fig:mechanism}--\ref{fig:heatmap}. Layer indices
are zero-based to match the figure annotations.

\begin{table}[t]
  \centering
  \small
  \setlength{\tabcolsep}{5pt}
  \begin{tabular}{@{}llcc@{}}
    \toprule
    \textbf{Model} & \textbf{Dataset} & $\ell^\ast$ & \textbf{AUROC} \\
    \midrule
    Qwen2.5-7B     & HaluEval         & L21         & 0.6332         \\
    Qwen2.5-7B     & SQuAD2           & L4          & 0.7329         \\
    Qwen2.5-7B     & HotpotQA         & L27         & 0.7603         \\
    Qwen2.5-7B     & TriviaQA         & L27         & 0.7199         \\
    \midrule
    Llama-3-8B     & HaluEval         & L10         & 0.7274         \\
    Llama-3-8B     & SQuAD2           & L11         & 0.7541         \\
    Llama-3-8B     & HotpotQA         & L31         & 0.8638         \\
    Llama-3-8B     & TriviaQA         & L31         & 0.8340         \\
    \midrule
    Gemma-2-9B     & HaluEval         & L1          & 0.6994         \\
    Gemma-2-9B     & SQuAD2           & L2          & 0.7491         \\
    Gemma-2-9B     & HotpotQA         & L40         & 0.7881         \\
    Gemma-2-9B     & TriviaQA         & L40         & 0.8033         \\
    \bottomrule
  \end{tabular}
  \caption{Peak-discriminative layer $\ell^\ast$ and the corresponding
    single-layer AUROC for $H_x^\ell$.}
  \label{tab:peak}
\end{table}

\begin{figure}[t]
  \centering
  \includegraphics[width=\columnwidth]{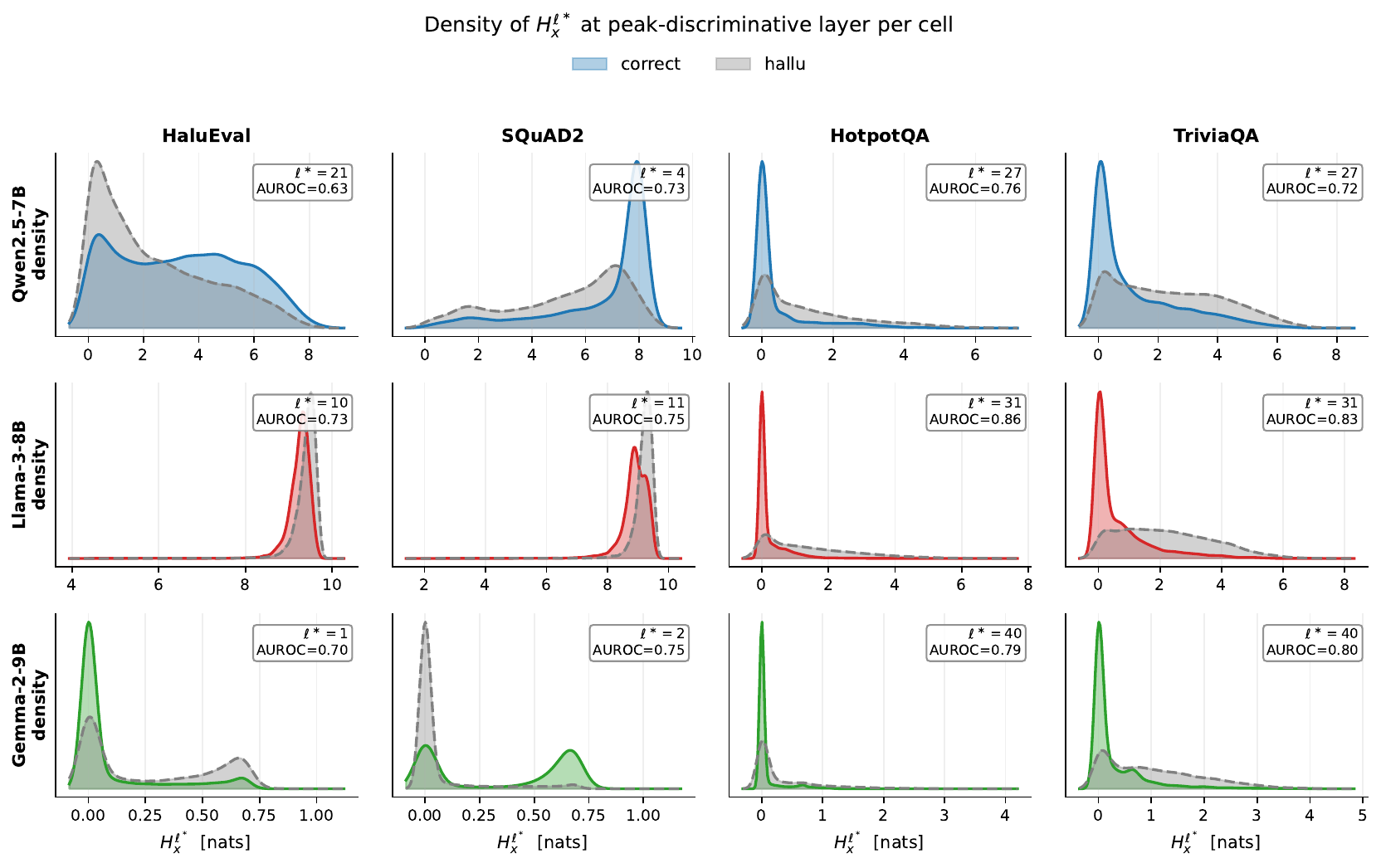}
  \caption{Probability density of $H_x^{\ell^\ast}$ at the
  peak-discriminative layer for correct vs.\ hallucinated responses.}
  \label{fig:density}
\end{figure}

\section{DoLa Defense: Full Configuration Grid}
\label{app:dola}

Table~\ref{tab:dola-full} provides the full configuration grid behind
Table~\ref{tab:dola}: both probe families and all four datasets, using
the same feature configuration as the main paper.

\begin{table}[t]
  \centering
  \small
  \setlength{\tabcolsep}{3pt}
  \resizebox{\columnwidth}{!}{%
    \begin{tabular}{@{}l|lccc@{}}
      \toprule
      \textbf{Probe} & \textbf{Dataset} & \textbf{DoLa-JSD} & \textbf{TriLens} & \textbf{TriLens+DoLa} \\
      \midrule
      \multirow{4}{*}{MLP}
                     & HaluEval         & 0.7623            & 0.9136           & \textbf{0.9155}       \\
                     & SQuAD2           & 0.8538            & 0.9158           & \textbf{0.9172}       \\
                     & HotpotQA         & 0.8928            & \textbf{0.9425}  & 0.9422                \\
                     & TriviaQA         & 0.8679            & \textbf{0.8861}  & 0.8860                \\
      \midrule
      \multirow{4}{*}{Linear}
                     & HaluEval         & 0.6790            & 0.8731           & \textbf{0.8738}       \\
                     & SQuAD2           & 0.7189            & 0.8778           & \textbf{0.8798}       \\
                     & HotpotQA         & 0.8111            & 0.9253           & \textbf{0.9256}       \\
                     & TriviaQA         & 0.8383            & 0.8700           & \textbf{0.8710}       \\
      \bottomrule
    \end{tabular}
  }
  \caption{Full DoLa comparison grid under both probe families.}
  \label{tab:dola-full}
\end{table}

\section{Efficiency and Complexity Analysis}
\label{app:hparams}

\subsection{Training Hyperparameters}

\begin{table}[t]
  \centering
  \small
  \begin{tabular}{@{}p{0.46\columnwidth}p{0.44\columnwidth}@{}}
    \toprule
    Model weights dtype         & bfloat16                                           \\
    Logit-lens projection dtype & float32                                            \\
    Attention implementation    & SDPA                                               \\
    Temperature $\tau$          & 1.0                                                \\
    Max response tokens         & 32                                                 \\
    Train/test split            & 80/20 group-preserving split                       \\
    Seeds                       & 5                                                  \\
    Main-paper dataset sizes    & reported in Section~\ref{sec:exp-setup}            \\
    Row labels per dataset      & exactly balanced correct/hallucinated labels       \\
    Linear probe                & L2 logreg, $C=1$                                   \\
    MLP probe architecture      & $3L \to 128 \to 64 \to 32 \to 1$                   \\
    MLP activation              & LeakyReLU(0.01)                                    \\
    MLP regularization          & BatchNorm + Dropout 0.3                            \\
    MLP optimizer               & Adam, lr $=5\times10^{-4}$                         \\
    MLP training                & 50 epochs, batch 32                                \\
    MLP LR schedule             & ReduceLROnPlateau ($\text{p}{=}5, \text{f}{=}0.5$) \\
    \bottomrule
  \end{tabular}
  \caption{Training hyperparameters used in the main experiments.}
  \label{tab:hparams-appendix}
\end{table}

\subsection{Time Complexity}

For a decoder with $L$ layers and hidden size $d$, TriLens adds a
logit-lens readout at three locations per layer and reduces each
readout to a scalar entropy. Relative to methods that store or classify
high-dimensional hidden states, the resulting sample representation is
compact: TriLens produces a $3L$-dimensional vector, whereas hidden-state
baselines typically operate on features that scale with $d$ or selected
layer subsets of size comparable to $d$. Probe-time complexity is
therefore negligible compared with the upstream LLM forward pass; in
practice, extraction dominates runtime and the downstream classifier is
cheap to retrain across seeds.

\subsection{Memory Footprint of Logit-Lens Extraction}

For a model with vocabulary size $|V|$, a naive implementation could
materialize, for each scored token, $3L$ intermediate logit-lens
distributions of the form
$\mathrm{softmax}(W_U \cdot \mathrm{Norm}(z)) \in \mathbb{R}^{|V|}$.
In float32, this corresponds to a peak memory cost of
$3L|V| \times 4$ bytes per token if all such distributions are retained
simultaneously.
For example, for Llama-3-8B ($L{=}32$, $|V|{\approx}128\mathrm{K}$),
the raw peak is approximately
$3 \times 32 \times 128000 \times 4 \approx 49$ MB per token.
In our implementation, however, each distribution is used immediately
to compute its Shannon entropy and is then discarded, so the persistent
storage scales as $O(3L)$ rather than $O(3L|V|)$.
For Llama-3-8B, this reduces the retained per-token state to only
96 scalar entropy values.

\end{document}